\documentclass[11pt,letterpaper]{article}

\usepackage[utf8]{inputenc}
\usepackage[T1]{fontenc}
\usepackage{lmodern}
\usepackage{microtype}
\usepackage[margin=1in]{geometry}
\usepackage{graphicx}
\usepackage{amsmath, amssymb}
\usepackage{booktabs}
\usepackage{multirow}
\usepackage{array}
\usepackage{caption}
\usepackage{hyperref}
\usepackage{xcolor}
\usepackage{enumitem}
\usepackage{authblk}

\hypersetup{
    colorlinks=true,
    linkcolor=blue,
    citecolor=blue,
    urlcolor=blue,
}

\title{\textbf{Sparse-LiDAR Prompting of Monocular Geometry Foundations:\\An Empirical Study Toward Long-Range Driving Depth}}

\author[1]{Kai Zheng}
\author[1]{Qiang Feng}
\author[1]{Xingjian Liu}
\author[1]{Wenquan Tan}
\author[1]{Yuan Li}
\affil[1]{Benewake (Beijing) Co., Ltd.}

\date{}

\begin{document}

\maketitle

\begin{abstract}
\noindent
Sparse-LiDAR-prompted depth foundation models (PromptDA, Prior Depth Anything, DMD3C) have shown strong results on indoor scenes or within KITTI's standard 80-meter evaluation cap. However, two limitations remain: (i) systematic distance-stratified evaluation in long-range driving regimes (50--150~m) is largely absent; (ii) prior approaches built on disparity-based foundations rely on pre-interpolated dense priors, leaving truly sparse LiDAR injection on point-map foundations (e.g., MoGe-2, NeurIPS 2025) unexplored. We present \textbf{SLIM} (Sparse-LiDAR Injected Monocular geometry), the first adaptation of MoGe-2 to accept truly sparse LiDAR input. SLIM integrates a partial-convolution sparse encoder with a multi-scale fusion neck that fuses LiDAR features into the point-map decoder at five scales. We adopt density-agnostic training (random injection ratio in $[0.005, 0.30]$) so a single model serves diverse input densities. On Virtual KITTI and CARLA, SLIM reduces the absolute relative error of the MoGe-2 baseline by approximately 39--51\% at 100--150~m. Ablation across six injection ratios shows partial-convolution injection improves both AbsRel and RMSE on Virtual KITTI in all six settings; on CARLA, AbsRel improves in five of six settings (one near-tie at 0.015 differs by 0.0013), and RMSE is comparable across encoders, with partial-convolution improving in three settings (by up to 0.31~unit) and losing by at most 0.11~unit in the other three.
\end{abstract}

\section{Introduction}
\label{sec:intro}

Reliable metric depth at 50--150~m is critical for autonomous driving. Yet recent sparse-LiDAR-prompted depth foundation models---PromptDA~\cite{lin2024promptda}, Prior Depth Anything~\cite{prior2025}, DMD3C~\cite{liang2025dmd3c}---have been primarily validated either on indoor benchmarks at sub-10-meter depths or within KITTI Depth Completion's 80-meter evaluation cap. To our knowledge, no prior work reports distance-stratified evaluation of sparse-LiDAR-prompted depth foundation models in driving scenarios beyond 80~m.

A second observation: existing approaches typically build on \emph{disparity-based} foundations~\cite{yang2024depthanything} and pre-interpolate sparse LiDAR into dense maps before fusion (PromptDA's design). In contrast, \emph{point-map foundations} (e.g., MoGe-2~\cite{wang2025moge2}) regress 3D point maps directly with metric scale but are monocular-only by design. Two consequences follow: (i) point-map foundations remain unexplored as a base for sparse-LiDAR prompting; (ii) pre-interpolation discards the binary nature of sparse projections.

We address both gaps with a partial-convolution~\cite{liu2018partialconv} sparse encoder and a five-scale fusion neck. Training proceeds under a randomly sampled injection ratio in $[0.005, 0.30]$, yielding density-agnostic adaptation.

\paragraph{Contributions.}
\textbf{(C1)} First distance-stratified evaluation of sparse-LiDAR-prompted depth foundation models in driving (50--150~m). \textbf{(C2)} First adaptation of MoGe-2 to accept truly sparse LiDAR input, in contrast to pre-interpolated dense priors used by prior frameworks. \textbf{(C3)} Density-agnostic training: a single model spans the $[0.005, 0.30]$ injection-ratio regime.

\section{Method}
\label{sec:method}

\begin{figure}[t]
    \centering
    \includegraphics[width=0.85\textwidth]{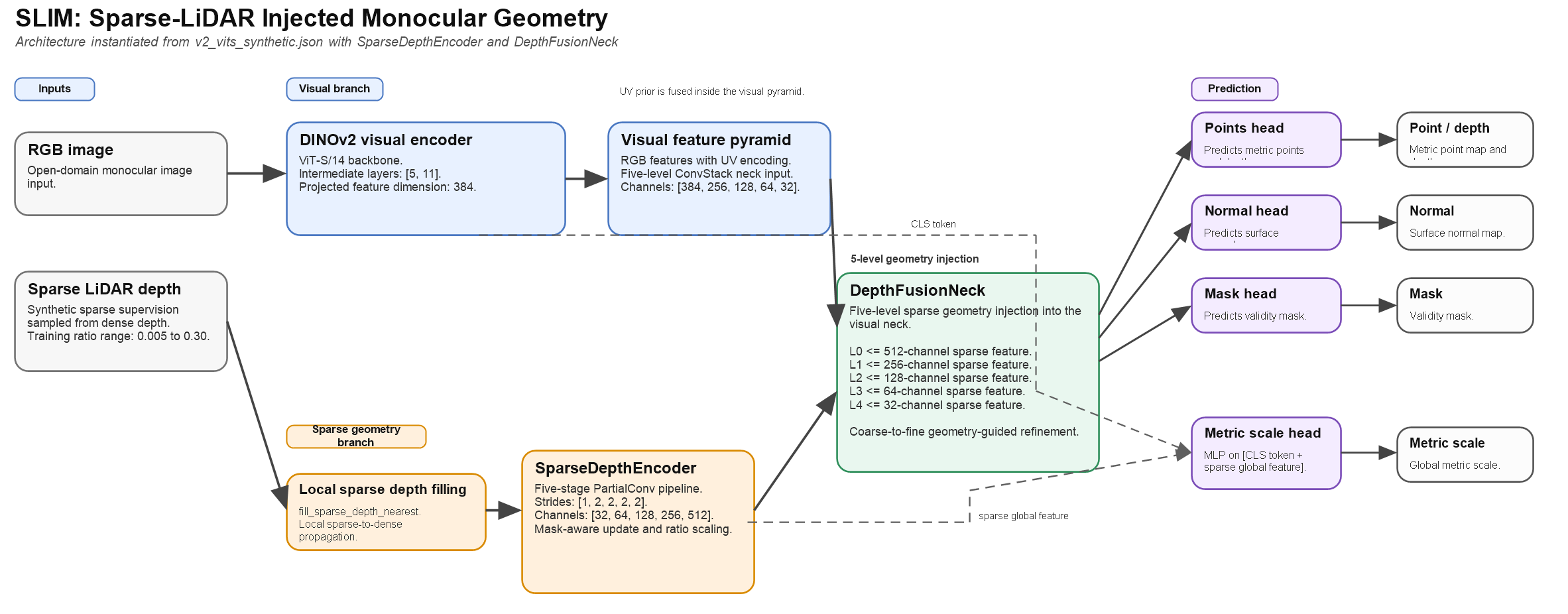}
    \caption{\textbf{SLIM architecture.} The visual branch is the MoGe-2~\cite{wang2025moge2} backbone (DINOv2 ViT-S/14; intermediate features taken from layers 5 and 11; projected feature dimension 384), producing a five-level visual feature pyramid (channels 384, 256, 128, 64, 32). The sparse-geometry branch applies local sparse-depth filling (nearest-neighbor propagation) followed by a five-stage PartialConv encoder (strides $[1, 2, 2, 2, 2]$; channels $[32, 64, 128, 256, 512]$) with mask-aware updates and ratio scaling. The DepthFusionNeck injects sparse-geometry features into the visual pyramid at five levels (L0--L4) for coarse-to-fine geometry-guided refinement. The decoder produces the metric 3D point map; auxiliary MoGe-2 heads predict surface normal, validity mask, and global metric scale (the latter via an MLP on the concatenation of the CLS token and a global sparse feature).}
    \label{fig:architecture}
\end{figure}

We extend MoGe-2~\cite{wang2025moge2} with three components, illustrated in Figure~\ref{fig:architecture}: a partial-convolution sparse encoder, a five-scale fusion neck, and a density-agnostic training procedure.

\paragraph{Sparse Depth Encoder.}
Sparse LiDAR projections (with valid pixel ratio sampled uniformly in $[0.005, 0.30]$ during training) are processed by partial convolutions~\cite{liu2018partialconv}: each layer applies a masked convolution normalized by the valid-pixel count within its receptive field, and the validity mask is propagated to subsequent layers. Five feature maps are produced at channel widths $[32, 64, 128, 256, 512]$.

\paragraph{Multi-Scale Fusion Neck.}
The fusion neck replaces MoGe-2's monocular ConvStack neck. Each FeatureFusionBlock combines RGB and depth features as
\begin{equation}
F_\text{fused} = F_\text{rgb} + \text{SE}(\text{DWConv}(\text{Conv}([F_\text{rgb}; F_\text{depth}]))),
\end{equation}
with SE channel attention~\cite{hu2018senet} (reduction ratio $r=4$) and a post-Conv batch normalization initialized with $\gamma=0.01$. The residual connection ensures near-identity behavior at the start of training, enabling stable fine-tuning from the MoGe-2 checkpoint.

\paragraph{Density-Agnostic Training.}
The injection ratio is sampled uniformly in $[0.005, 0.30]$ at each training step. Auxiliary loss terms include edge-aware reweighting based on GT 3D-neighbor distance, log-distance weighting in lieu of conventional $1/z$, and a sparse-depth consistency term enforced at valid LiDAR pixels.

\section{Experiments}
\label{sec:experiments}

\paragraph{Setup.}
We evaluate on Virtual KITTI 2 (vk; $\sim$21K frames)~\cite{cabon2020vk2} and CARLA (ca)~\cite{dosovitskiy2017carla}; the CARLA scenes are augmented with long-range small targets to probe the regime of interest, with identical augmentation applied to all evaluated models. The backbone is MoGe-2 ViT-S/14 (35.1M parameters). We use the term \emph{injection ratio} (synonymous with \emph{sparsity} in our usage) to denote the fraction of pixels carrying valid LiDAR depth; lower values correspond to sparser inputs. Evaluation is reported in three distance bins (1--50, 50--100, 100--150~m) using AbsRel, RMSE, and $\delta < 1.25$.

\subsection{Distance-Stratified Results}
\label{sec:distance}

\begin{table}[t]
\centering
\small
\caption{\textbf{Distance-stratified results.} SLIM vs. the MoGe-2 monocular baseline at injection ratio 0.005. Bold indicates the better value per metric.}
\label{tab:distance}
\begin{tabular}{lcccccccc}
\toprule
\textbf{Range} & \textbf{Model} & AbsRel(vk) & AbsRel(ca) & RMSE(vk) & RMSE(ca) & $\delta_1$(vk) & $\delta_1$(ca) \\
\midrule
\multirow{2}{*}{1--50~m}
& \textbf{SLIM}   & \textbf{0.071} & \textbf{0.095} & \textbf{4.56} & 6.14 & \textbf{0.951} & \textbf{0.953} \\
& MoGe-2          & 0.178 & 0.111 & 5.02 & \textbf{4.08} & 0.692 & 0.894 \\
\midrule
\multirow{2}{*}{50--100~m}
& \textbf{SLIM}   & \textbf{0.097} & \textbf{0.091} & \textbf{13.93} & \textbf{15.70} & \textbf{0.873} & \textbf{0.893} \\
& MoGe-2          & 0.212 & 0.180 & 18.54 & 16.34 & 0.645 & 0.749 \\
\midrule
\multirow{2}{*}{\textbf{100--150~m}}
& \textbf{SLIM}   & \textbf{0.133} & \textbf{0.118} & \textbf{27.09} & \textbf{25.94} & \textbf{0.798} & \textbf{0.823} \\
& MoGe-2          & 0.270 & 0.194 & 37.75 & 29.38 & 0.493 & 0.676 \\
\midrule
\multirow{2}{*}{Overall}
& \textbf{SLIM}   & \textbf{0.074} & \textbf{0.097} & \textbf{11.87} & \textbf{12.89} & \textbf{0.940} & \textbf{0.931} \\
& MoGe-2          & 0.187 & 0.135 & 15.34 & 15.75 & 0.672 & 0.841 \\
\bottomrule
\end{tabular}
\end{table}

\begin{figure}[t]
    \centering
    \includegraphics[width=0.7\textwidth]{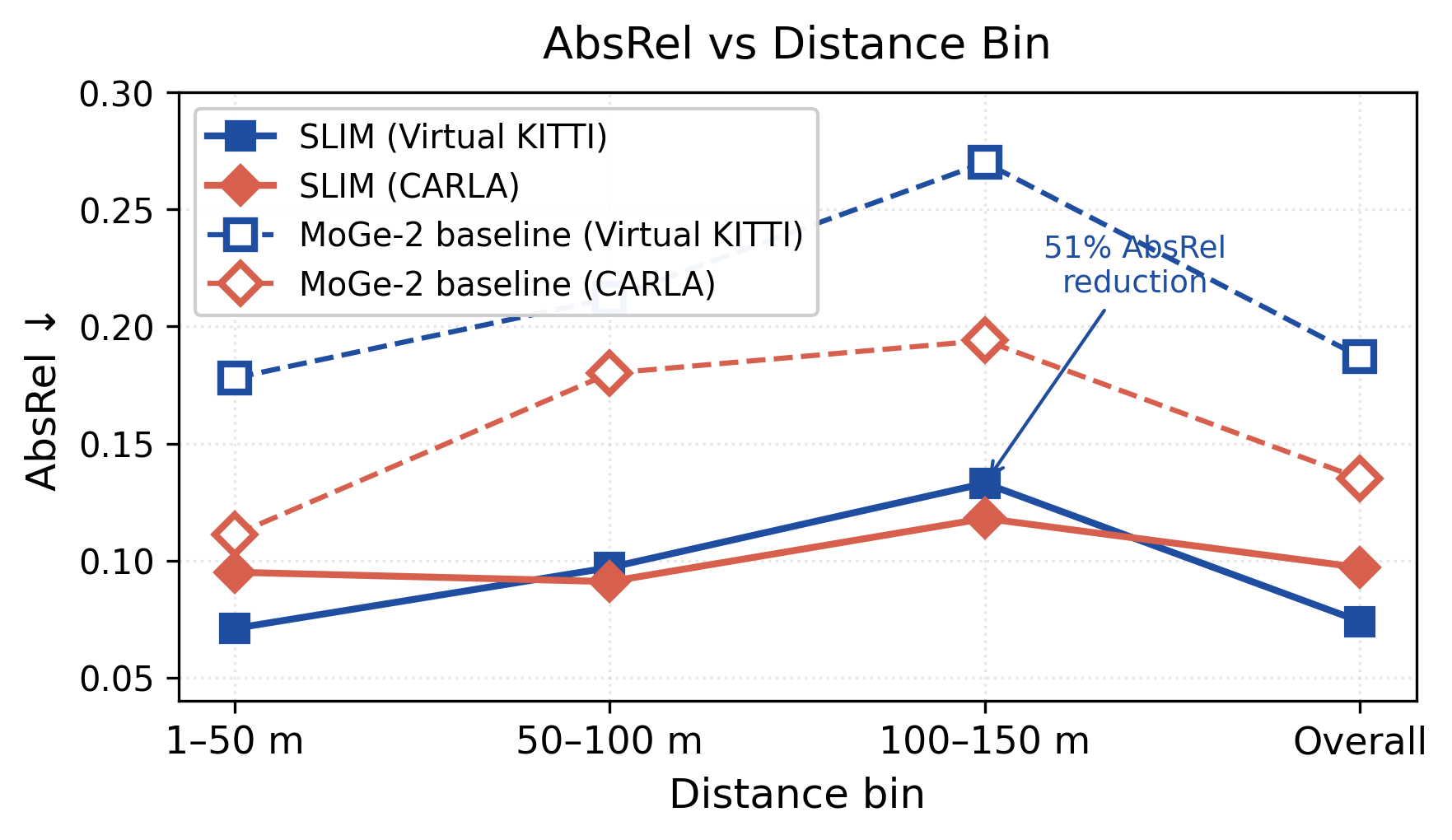}
    \caption{\textbf{AbsRel as a function of distance bin.} SLIM's curve remains comparatively flat across distance, whereas the MoGe-2 baseline's error grows steeply in the long-range regime. Drawn from Table~\ref{tab:distance}.}
    \label{fig:distance}
\end{figure}

At 100--150~m, SLIM reduces AbsRel by 50.8\% on Virtual KITTI and by 39.5\% on CARLA, relative to the MoGe-2 baseline. Across all three distance bins, the baseline's error grows steeply with distance, while SLIM's error remains comparatively stable.

We do not directly compare with PromptDA~\cite{lin2024promptda}, whose released architecture requires bilinear pre-interpolation of sparse depth as input. Instead, we implement a \emph{pre-interpolation baseline} (\S\ref{sec:ablation}) that follows the same design principle within our framework, enabling a controlled comparison of the sparse-encoding choice in isolation.

\subsection{PartialConv vs.\ Pre-Interpolation Ablation}
\label{sec:ablation}

\begin{table}[t]
\centering
\small
\caption{\textbf{Sparse encoder ablation across six injection ratios.} Bold indicates the better value per metric.}
\label{tab:ablation}
\begin{tabular}{cccccc}
\toprule
\textbf{Inj.} & \textbf{Encoder} & AbsRel(vk) & AbsRel(ca) & RMSE(vk) & RMSE(ca) \\
\midrule
\multirow{2}{*}{0.005}
& \textbf{PartialConv}    & \textbf{0.074} & \textbf{0.097} & \textbf{11.87} & \textbf{12.89} \\
& Interpolation           & 0.077 & 0.115 & 11.96 & 13.20 \\
\midrule
\multirow{2}{*}{0.008}
& \textbf{PartialConv}    & \textbf{0.065} & \textbf{0.101} & \textbf{10.87} & \textbf{12.09} \\
& Interpolation           & 0.068 & 0.104 & 10.97 & 12.22 \\
\midrule
\multirow{2}{*}{0.010}
& \textbf{PartialConv}    & \textbf{0.062} & \textbf{0.095} & \textbf{10.39} & 11.74 \\
& Interpolation           & 0.065 & 0.099 & 10.45 & \textbf{11.73} \\
\midrule
\multirow{2}{*}{0.015}
& \textbf{PartialConv}    & \textbf{0.058} & 0.095 & \textbf{9.67} & 11.16 \\
& Interpolation           & 0.063 & \textbf{0.094} & 9.75 & \textbf{11.05} \\
\midrule
\multirow{2}{*}{0.020}
& \textbf{PartialConv}    & \textbf{0.056} & \textbf{0.087} & \textbf{9.19} & \textbf{10.29} \\
& Interpolation           & 0.061 & 0.090 & 9.28 & 10.37 \\
\midrule
\multirow{2}{*}{0.030}
& \textbf{PartialConv}    & \textbf{0.054} & \textbf{0.083} & \textbf{8.52} & 9.68 \\
& Interpolation           & 0.060 & 0.085 & 8.60 & \textbf{9.67} \\
\bottomrule
\end{tabular}
\end{table}

Across the six tested injection ratios, partial-convolution injection improves both AbsRel and RMSE on Virtual KITTI in every setting. On CARLA, AbsRel improves in five of six settings (the remaining setting at 0.015 is a near-tie, differing by 0.0013). RMSE on CARLA is comparable across encoders: partial-convolution wins in three of six settings (by 0.08--0.31~unit) and loses by at most 0.11~unit in the other three. The largest improvement appears at the sparsest setting (0.005), where pre-interpolation introduces the most spurious dense values.

\subsection{Qualitative}
\label{sec:qualitative}

\begin{figure}[t]
    \centering
    \includegraphics[width=0.85\textwidth]{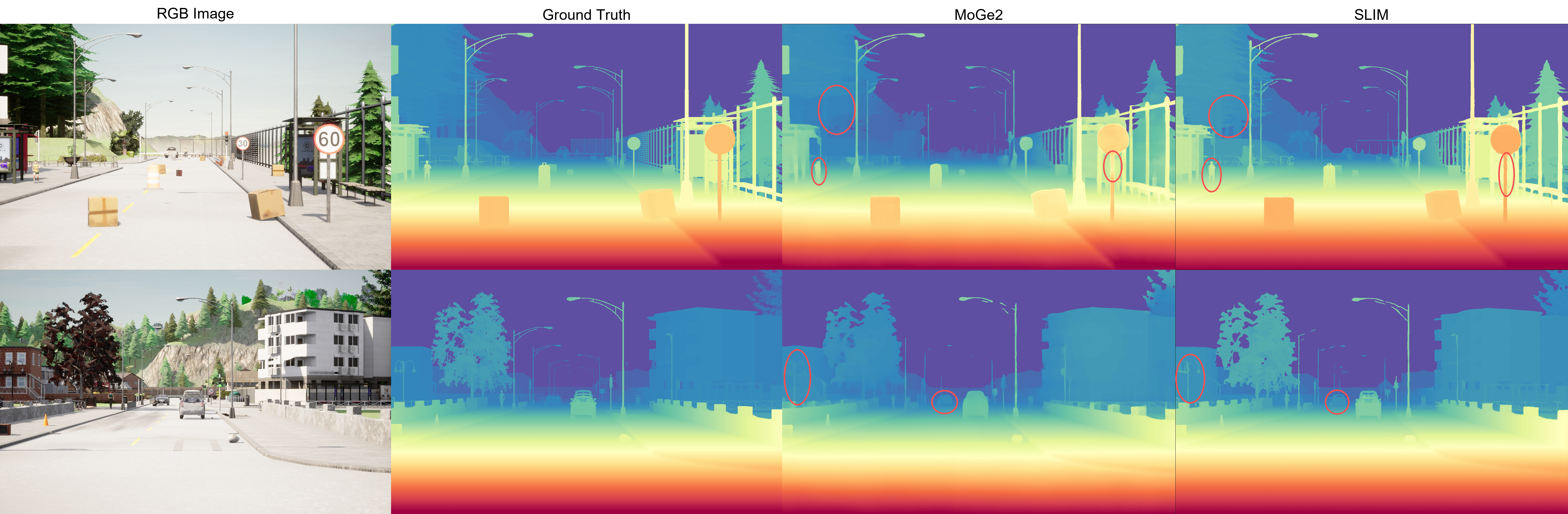}
    \caption{\textbf{Qualitative comparison on two Virtual KITTI scenes.} Four columns from left to right: RGB image, ground-truth depth, MoGe-2 monocular baseline, and SLIM (ours). Red circles mark regions where the baseline exhibits errors in fine or distant structures.}
    \label{fig:qualitative}
\end{figure}

In the first row, the MoGe-2 baseline produces blurred depth textures for the background trees (top-left circle), renders the pedestrian on the sidewalk barely recognizable (bottom-left circle), and gives inconsistent depth estimates for the traffic sign (right circle). SLIM maintains sharper and more consistent depth across all three regions: distant trees retain clearer texture, the pedestrian's silhouette is resolved, and the traffic sign aligns more closely with the ground truth.

In the second row, MoGe-2 again struggles with fine and distant structures: the distant streetlight pole becomes nearly indistinguishable from the background (left circle), and a small obstacle in front of the vehicle merges into the car's depth map, losing its distinct structure (center circle). SLIM preserves clear, well-defined depth for both objects.

\section{Discussion, Limitations, and Conclusion}
\label{sec:conclusion}

\paragraph{V2X cooperative perception.}
The growing monocular error in the long-range regime motivates physical anchoring via sparse LiDAR. A vehicle equipped with sparse LiDAR can serve as a physical-anchor node, sharing anchor points with neighboring vehicles or roadside units that run monocular geometry foundations.

\paragraph{Practical deployment.}
Density-agnostic training obviates sensor-specific retraining across the $[0.005, 0.30]$ injection-ratio regime. The 35.1M-parameter footprint is compatible with onboard deployment; a comprehensive latency study is left to future work.

\paragraph{Limitations.}
Our quantitative claims are limited to Virtual KITTI and CARLA. Public benchmark evaluation---KITTI Depth Completion against DMD3C and Prior Depth Anything, nuScenes, and DDAD---is critical future work. Real-world transfer under motion blur, weather, and material-induced multipath remains untested. Industrial scanning patterns (line-scan, foveated FOV) warrant pattern-aware training. Integration with dynamic-aware ground-truth generation pipelines~\cite{docdepth2025} for moving-object regions and test-time adaptation methods~\cite{ttadepth2024} are complementary directions for future work.

\paragraph{Conclusion.}
We presented SLIM, the first adaptation of MoGe-2 to truly sparse LiDAR input, and the first distance-stratified evaluation of sparse-LiDAR-prompted depth foundation models in long-range driving (50--150~m). On synthetic datasets, SLIM reduces AbsRel by approximately 39--51\% at 100--150~m relative to the MoGe-2 baseline. We hope this study motivates broader adoption of distance-stratified evaluation and continued exploration of point-map foundations for long-range driving perception.


\end{document}